\documentclass[10pt,twocolumn,letterpaper]{article}

\usepackage{cvpr}              % To produce the CAMERA-READY version
\usepackage[accsupp]{axessibility}
\usepackage{graphicx}
\usepackage{amsmath}
\usepackage{amssymb}
\usepackage{booktabs}
\usepackage{color, colortbl}
\usepackage{dsfont}
\usepackage{microtype}
\usepackage{threeparttable}
\usepackage{soul}
\usepackage{multirow}

\usepackage{graphicx}
\usepackage{amsmath}
\usepackage{amssymb}
\usepackage{booktabs}
\usepackage{tabularx} % in the preamble
\usepackage{enumitem}
\usepackage{arydshln}
\usepackage{algorithm}
\usepackage[noend]{algpseudocode}
\usepackage{wrapfig}

\usepackage{mwe}
\long\def\/*#1*/{}
\usepackage{pifont}% http://ctan.org/pkg/pifont
\usepackage{times}
\usepackage{dsfont}
\usepackage{epsfig}
\usepackage{microtype}
\usepackage{cite}
\usepackage{color, colortbl}
\usepackage{comment}
\usepackage{multirow}
\usepackage{adjustbox}
\usepackage{wrapfig}

\def\eg{\emph{e.g}\onedot} 
\def\ie{\emph{i.e}\onedot}

\def\etal{\emph{et al}\onedot}

\newcommand{\xmark}{\ding{55}}%
\usepackage{booktabs}
\usepackage{microtype}
\usepackage{threeparttable}
\usepackage{soul}

\definecolor{Gray}{gray}{0.9}

\usepackage[pagebackref,breaklinks,colorlinks]{hyperref}

\usepackage[capitalize]{cleveref}
\crefname{section}{Sec.}{Secs.}
\Crefname{section}{Section}{Sections}
\Crefname{table}{Table}{Tables}
\crefname{table}{Tab.}{Tabs.}

\def\cvprPaperID{9219} % *** Enter the CVPR Paper ID here
\def\confName{CVPR}
\def\confYear{2022}

\begin{document}

\title{TubeR: Tubelet Transformer for Video Action Detection}
\author{Jiaojiao Zhao\textsuperscript{\rm 1}\thanks{Equally contributed and work done while at AWS AI Labs},  Yanyi Zhang\textsuperscript{\rm 2}\footnotemark[1], Xinyu Li\textsuperscript{\rm 3}\footnotemark[1], Hao Chen\textsuperscript{\rm 3}, Bing Shuai\textsuperscript{\rm 3}, Mingze Xu\textsuperscript{\rm 3}, Chunhui Liu\textsuperscript{\rm 3},\\Kaustav Kundu\textsuperscript{\rm 3}, Yuanjun Xiong\textsuperscript{\rm 3}, Davide Modolo\textsuperscript{\rm 3}, Ivan Marsic\textsuperscript{\rm 2}, Cees G.M. Snoek\textsuperscript{\rm 1}, Joseph Tighe\textsuperscript{\rm 3}\\
\textsuperscript{\rm 1}University of Amsterdam \hspace{2mm} \textsuperscript{\rm 2}Rutgers University\hspace{2mm} \textsuperscript{\rm 3}AWS AI Labs\\
%{\tt\small \{j.zhao3, cgmsnoek\}@uva.nl; \{yz593, marsic\}@rutgers.edu;}\\ \tt\small {\{xxnl,hxen,bshuai,kaustavk,xumingze,chunhliu,yuanjx,dmodolo,tighej\}@amazon.com}
}

\maketitle

%%%%%%%%% ABSTRACT
\begin{abstract}
We propose TubeR: a simple solution for spatio-temporal video action detection. Different from existing methods that depend on either an off-line actor detector or hand-designed actor-positional hypotheses like proposals or anchors,
we propose to directly detect an action tubelet in a video by simultaneously performing action localization and recognition from a single representation. TubeR learns a set of tubelet-queries and utilizes a tubelet-attention module to model the dynamic spatio-temporal nature of a video clip, which effectively reinforces the model capacity compared to using actor-positional hypotheses in the spatio-temporal space. For videos containing transitional states or scene changes, we propose a context aware classification head to utilize short-term and long-term context to strengthen action classification, and an action switch regression head for detecting the precise temporal action extent. TubeR directly produces action tubelets with variable lengths and even maintains good results for long video clips.
TubeR outperforms the previous state-of-the-art on commonly used action detection datasets AVA, UCF101-24 and JHMDB51-21. Code will be available on GluonCV(https://cv.gluon.ai/). 
\end{abstract}

%%%%%%%%% 
\section{Introduction}
\label{sec:intro}

\begin{figure}[t]
    \centering
    \includegraphics[width=0.45\textwidth]{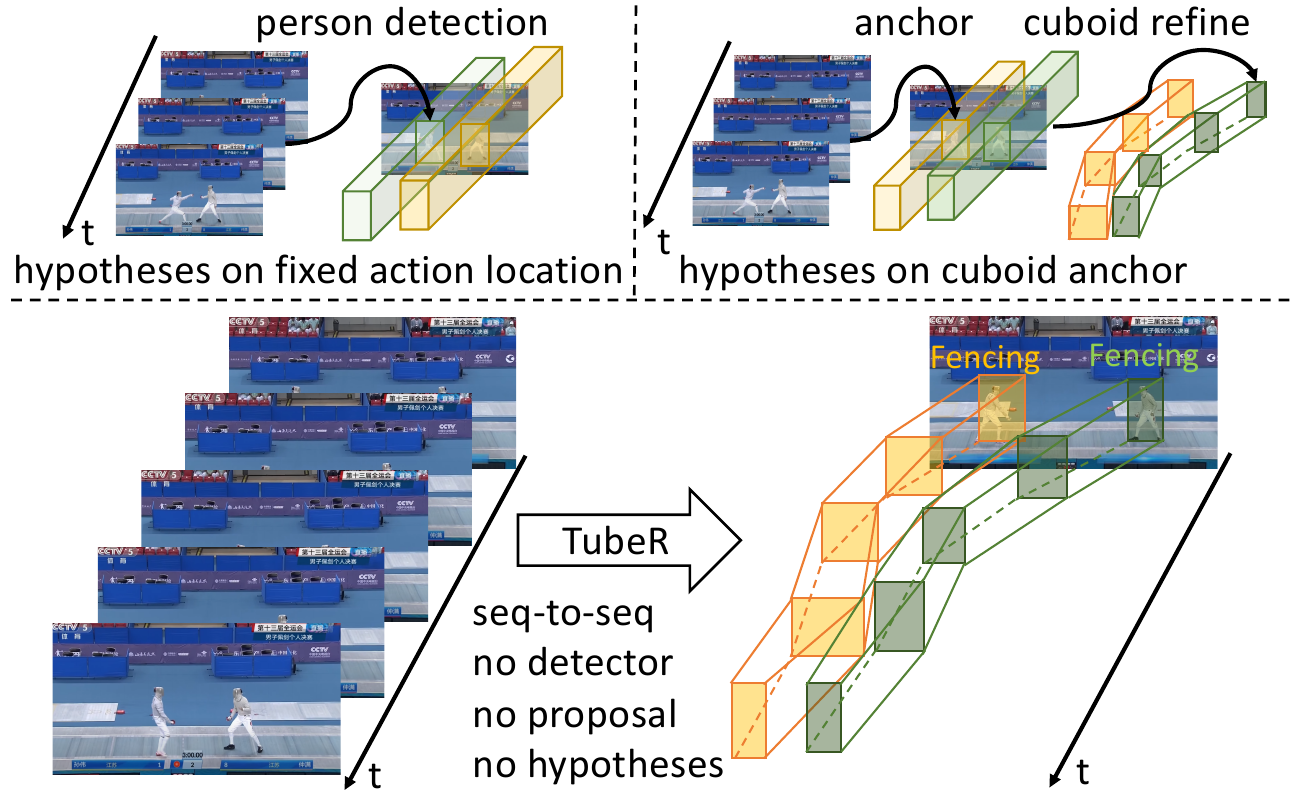}
    \caption{TubeR takes as input a video clip and directly outputs tubelets: sequences of bounding boxes and their action labels. TubeR runs end-to-end without person detectors, anchors or proposals.}
    \label{fig:teaser}
    \vspace{-5mm}
\end{figure}

This paper tackles the problem of spatio-temporal human action detection in videos~\cite{cao2010cross,tran2012max,jain2014action}, which plays a central role in advanced video search engines, robotics, and self-driving cars. Action detection is a compound task, requiring the localization of per-frame person instances, the linking of these detected person instances into action tubes and the prediction of their action class labels. Two approaches for spatio-temporal action detection are prevalent in the literature: frame-level detection and tubelet-level detection. Frame-level detection approaches detect and classify the action independently on each frame\cite{gkioxari2015finding,peng2016multi,singh2017online}, and then link per-frame detections together into coherent action tubes. To compensate for the lack of temporal information, several methods simply repeat 2D proposals~\cite{gu2018ava,sun2018actor,girdhar2018better} or offline person detections~\cite{feichtenhofer2019slowfast,wu2019long,tang2020asynchronous,pan2021actor} over time to obtain spatio-temporal features (Figure \ref{fig:teaser} top left). 
%
%====================
%===================

Alternatively, tubelet-level detection approaches~\cite{hou2017tube,kalogeiton2017action,zhao2019dance,song2019tacnet,yang2019step,li2020actions}, directly generate spatio-temporal volumes from a video clip to capture the coherence and dynamic nature of actions. 
They typically predict action localization and classification jointly over spatio-temporal hypotheses, like 3D cuboid proposals~\cite{hou2017tube,kalogeiton2017action}  (Figure \ref{fig:teaser} top right). 
Unfortunately, these 3D cuboids can only capture a short period of time, also when the spatial location of a person changes as soon as they move, or due to camera motion. Ideally, this family of models would use flexible spatio-temporal tubelets that can track the person over a longer time, but the large configuration space of such a parameterization has restricted previous methods to short cuboids. In this work we present a tubelet-level detection approach that is able to simultaneously localize and recognize action tubelets in a flexible manner, which allows tubelets to change in size and location over time (Figure \ref{fig:teaser} bottom). This allows our system to leverage longer tubelets, which aggregate visual information of a person and their actions over longer periods of time.

We draw inspiration from sequence-to-sequence modelling in natural language processing (NLP), particularly machine translation~\cite{vaswani2017attention,li2019entangled,sur2020self,khan2020mmft}, and its application to object detection, DETR~\cite{carion2020end}. Being a detection framework, DETR can be applied as a frame-level action detection approach trivially, but the power of the transformer framework, on which DETR is built, is its ability to generate complex structured outputs over sequences. In NLP, this typically takes the form of sentences but in this work we use the notion of decoder queries to represent people and their actions over video sequences, without having to restrict tubelets to fixed cuboids.

We propose a tubelet-transformer, we call TubeR, for localizing and recognizing actions from a single representation. Building on the DETR framework~\cite{carion2020end}, TubeR learns a set of tubelet queries to pull action-specific tubelet-level features from a spatio-temporal video representation. Our TubeR design includes a specialized spatial and temporal tubelet attention to allow our tubelets to be unrestricted in their spatial location and scale over time, thus overcoming previous limitations of methods restricted to cuboids. TubeR regresses bounding boxes within a tubelet jointly across time, considering temporal correlations between tubelets, and aggregates visual features over the tubelet to classify actions. This core design already performs well, outperforming many previous model designs, but still does not improve upon frame-level approaches using offline person detectors. 
We hypothesize that this is partially due to the lack of more global context in our query based feature as it is hard to classify actions referring to relationships such as `listening-to' and `talking-to' by only looking at a single person.
Therefore, we introduce a context aware classification head that, along with the tubelet feature, takes the full clip feature from which our classification head can draw contextual information. This design allows the network to effectively relate a person tubelet to the full scene context where the tubelet appears and is shown to be effective on its own in our results section. One limitation of this design is the context feature is only drawn from the same clip our tubelet occupies. It has been shown~\cite{wu2019long} to be important to also include long term contextual features for the final action classification.
Thus, we introduce a memory system inspired by \cite{xu2021long} to compress and store contextual features from video content around the tubelet.
We feed this long term contextual memory to our classification head using the same feature injection strategy and again show this gives an important improvement over the short term context alone. We test our full system on three popular action detection datasets (AVA~\cite{gu2018ava}, UCF101-24~\cite{soomro2012ucf101} and JHMDB51-21~\cite{jhuang2013towards}) and show our method can outperform other state-of-the-art results.

In summary, our contributions are as follows:
\begin{enumerate}[itemsep=0pt,parsep=0pt]
    \item We propose TubeR: a tubelet-level transformer framework for human action detection.
    \item Our tubelet query and attention based formulation is able to generate tubelets of arbitrary location and scale.
    \item Our context aware classification head is able to aggregate short-term and long-term contextual information.
    \item We present state-of-the-art results on three challenging action detection datasets.
\end{enumerate}

%%%%%%%%% 
\section{Related Work}
\label{sec:relatedwork}

\noindent\textbf{Frame-level action detection.} Spatio-temporal action detection in video has a long tradition, \eg~\cite{cao2010cross,tran2012max,jain2014action,weinzaepfel2015learning,peng2016multi,gu2018ava,tang2020asynchronous,pan2021actor}. Inspired by object detection using deep convolution neural networks, action detection in video has been considerably improved by frame-level methods~\cite{weinzaepfel2015learning,peng2016multi,saha2016deep,singh2017online}. These methods perform localization and recognition per-frame and then link frame-wise predictions to action tubes. 
%
%Most recent methods handle localization and classification separately. 
%
Specifically, they apply 2D positional hypotheses (anchors) or an offline person detector on a keyframe for localizing actors, and then focus more on improving action recognition. They incorporate temporal patterns by an extra stream utilizing optical flow. Others~\cite{gu2018ava,girdhar2018better,sun2018actor} apply 3D convolution networks to capture temporal information for recognizing actions. Feichtenhofer~\etal~\cite{feichtenhofer2019slowfast} present a slowfast network to even better capture spatio-temporal information. Both Tang \etal~\cite{tang2020asynchronous} and Pan \etal~\cite{pan2021actor} propose to explicitly model relations between actors and objects. 
%There are more works~\cite{wu2019long,tang2020asynchronous,pan2021actor} follow this fashion to downgrade an action detection task to an action recognition task on the dataset AVA. 
Recently, Chen~\etal~\cite{chen2021watch} propose to train actor localization and action classification end-to-end from a single backbone. Different from these frame-level approaches, we target on tubelet-level video action detection, with a unified configuration to simultaneously perform localization and recognition.

\noindent\textbf{Tubelet-level action detection.} Detecting actions by taking a tubelet as a representation unit~\cite{li2018recurrent,zhao2019dance,song2019tacnet,yang2019step,li2020actions} has been popular since it was proposed by Jain~\etal~\cite{jain2014action}. Kalogeiton~\etal~\cite{kalogeiton2017action} repeat 2D anchors per-frame for pooling ROI features and then stack the frame-wise features to predict action labels. Hou~\etal\cite{hou2017tube} and Yang~\etal~\cite{yang2019step} depend on carefully-designed 3D cuboid proposals. The former directly detects tubelets and the later progressively refines 3D cuboid proposals across time. Besides box/cuboid anchors, Li~\etal~\cite{li2020actions} detect tubelet instances by relying on center position hypotheses. 
Hypotheses-based methods have difficulties to process long video clips, as we discussed in the introduction. 
We add to the tubelet tradition by learning a small set of tubelet queries to represent the dynamic nature of tubelets. We reformulate the action detection task as a sequence-to-sequence learning problem and explicitly model the temporal correlations within a tubelet. Our method is capable to handle long video clips.

\noindent\textbf{Transformer-based action detection.} Vaswani \etal \cite{vaswani2017attention} proposed the transformer for machine translation, which soon after became the most popular backbone for sequence-to-sequence tasks, \eg,~\cite{li2019entangled,sur2020self,khan2020mmft}. Recently, it has also demonstrated impressive advances in object detection~\cite{carion2020end,zhu2020deformable}, image classification~\cite{dosovitskiy2020image,yuan2021tokens} and video recognition ~\cite{gavrilyuk2020actor,zhang2021vidtr,fan2021multiscale}. Girdhar \etal \cite{girdhar2019video} propose a video action transformer network for detecting actions. They apply a region-proposal-network for localization. The transformer is utilized for further improving action recognition by aggregating features from the spatio-temporal context around actors. We propose a unified solution to simultaneously localize and recognize actions.

%BTW: Pengwan Yang, Pascal Mettes, Cees Snoek: Few-Shot Transformation of Common Actions into Time and Space. In: CVPR, 2021.
% --> Uses transformer for tubelet-based action detection in few-shot setting.

%%%%%%%%% 
\section{Action Detection by TubeR}
\label{sec:relatedwork}
\begin{figure}
    \centering
    \includegraphics[width=0.47\textwidth]{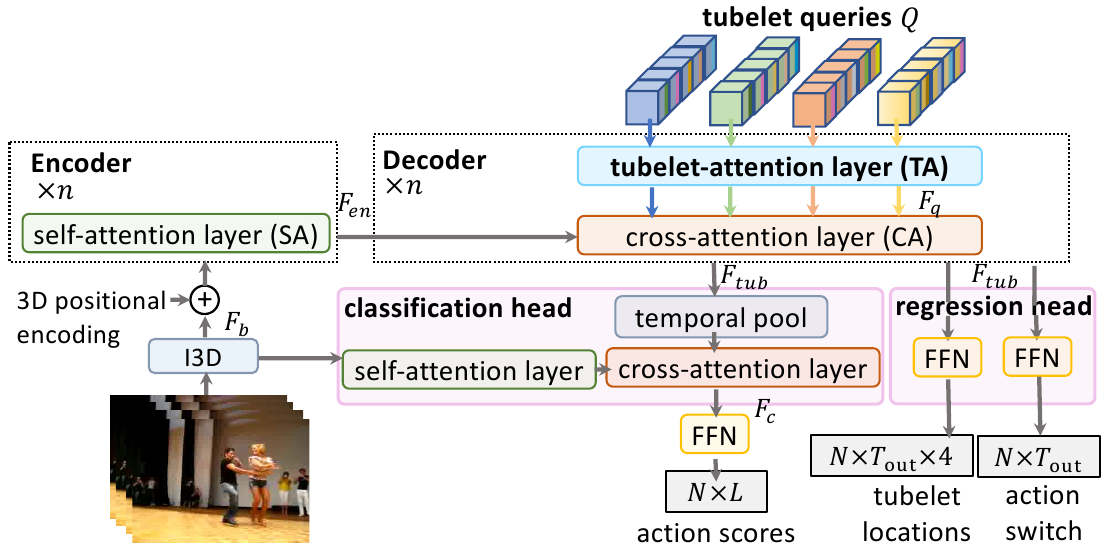}
    \caption{The overall structure of TubeR. Both encoder and decoder contain $n$ stacked modules. We only show the key components in the encoder and decoder modules. The encoder models the spatio-temporal features from the backbone $F_\text{b}$ by self-attention layers (see Section~\ref{sec:ed}). The decoder transforms a set of tubelet queries $Q$ and generates tubelet-level features $F_\text{tub}$. We utilize tubelet-attention layers to model the relations between box query embeddings within a tubelet (see Section~\ref{sec:query}). Finally, we apply the context aware classification head and action switch regression head to predict tubelet labels and coordinates (see Section~\ref{sec:head}).}
    \label{fig:overview}
    \vspace{-3mm}
\end{figure}

In this section, we present our TubeR that takes as input a video clip and directly outputs a tubelet: a sequence of bounding boxes and the action label. The TubeR design takes inspiration from the image-based DETR~\cite{carion2020end} but reformulates the transformer architecture for sequence-to-sequence(s) modeling in video (Figure~\ref{fig:overview}).

Given a video clip $I \in \mathbb{R}^{T_\text{in}\times H\times W\times C }$ where $T_\text{in},H,W,C$ denote the number of frames, height, width, and colour channels, TubeR first applies a 3D backbone to extract video feature $F_\text{b}\in \mathbb{R}^{T' \times H' \times W' \times C'}$, where $T'$ is the temporal dimension and $C'$ is the feature dimension. A transformer encoder-decoder is then utilized to transform the video feature into a set of tubelet-specific feature $F_{\text{tub}} \in \mathbb{R}^{N \times T_\text{out} \times C'}$, with $T_\text{out}$ the output temporal dimension and $N$ the number of tubelets. In order to process long video clips, we use temporal down-sampling to make $T_\text{out}<T'<T_\text{in}$, which reduces our memory requirement. In this case, TubeR generates sparse tubelets. For short video clips we remove the temporal down-sampling to make sure $T_\text{out}{=}T'{=}T_\text{in}$, which results in dense tubelets. Tubelet regression and associated action classification can be achieved simultaneously with separated task heads as:
\begin{equation}
    y_{\text{coor}} = f(F_{\text{tub}});
    y_{\text{class}} = g(F_{\text{tub}}),
    \label{equ:feature_head}
\end{equation}
where $f$ denotes the tubelet regression head and $y_{\text{coor}} \in \mathbb{R}^{N \times T_\text{out} \times 4}$ stands for the coordinates of $N$ tubelets, each of which is across $T_\text{out}$ frames (or $T_\text{out}$ sampled frames for long video clips). Here $g$ denotes the action classification head, and $y_{\text{class}} \in \mathbb{R}^{N \times L}$ stands for the action classification for $N$ tubelets with $L$ possible labels.

\subsection{TubeR Encoder}
\label{sec:ed}
Different from the vanilla transformer encoder, the TubeR encoder is designed for processing information in the 3D spatio-temporal space. Each encoder layer is made up of a self-attention layer (SA), two normalization layers and a feed forward network (FFN), following \cite{vaswani2017attention}. We only put the core attention layers in all equations below.
\begin{equation}
        F_{\text{en}} = \text{Encoder}(F_\text{b}), 
\end{equation}
\vspace{-5mm}
\begin{equation}
\label{equ:self_attn}
    \text{SA}(F_\text{b}) = \text{softmax}(\frac{\sigma_{q}(F_\text{b}) \times \sigma_{k}(F_\text{b})^T}{\sqrt{C'}})\times \sigma_{v}(F_\text{b}),
\end{equation}
\begin{equation}
    \sigma(*) = \text{Linear}(*) + \text{Emb}_{\text{pos}},
\end{equation}
where $F_\text{b}$ is the backbone feature and $F_{\text{en}} \in \mathbb{R}^{T'H'W' \times C'}$ denotes the $C'$ dimensional encoded feature embedding.  The $\sigma(*)$ is the linear transformation plus positional embedding.
$\text{Emb}_{\text{pos}}$ is the 3D positional embedding \cite{zhang2021vidtr}. The optional temporal down-sampling can be applied to the backbone feature to shrink the input sequence length to the transformer for better memory efficiency.

\subsection{TubeR Decoder}
\label{sec:query}

\noindent\textbf{Tubelet query.} Directly detecting tubelets is quite challenging based on anchor hypotheses. The tubelet space along the spatio-temporal dimension is huge compared to the single-frame bounding box space. Consider for example Faster-RCNN~\cite{ren2015faster} for object detection, which requires for each position in a feature map with spatial size $H' {\times} W'$, $K({=}9)$ anchors. There are in total $KH'W'$ anchors. For a tubelet across $T_\text{out}$ frames, it would require $(KH'W')^{T_\text{out}}$ anchors to maintain the same sampling in space-time. 
To reduce the tubelet space, several methods~\cite{hou2017tube,yang2019step} adopt 3D cuboids to approximate tubelets by ignoring the spatial action displacements in a short video clip. However, the longer the video clip is, the less accurately a 3D cuboid hypotheses represents a tubelet. We propose to learn a small set of tubelet queries $Q{=}\{Q_1,...,Q_{N}\}$ driven by the video data. $N$ is the number of queries. The $i$-th tubelet query $Q_i{=}\{{q_{i,1}, ..., q_{i,T_\text{out}}}\}$ contains $T_\text{out}$ box query embeddings $ q_{i,t}\in \mathbb{R}^{C'}$ across $T_\text{out}$ frames. We learn a tubelet query to represent the dynamics of a tubelet, instead of hand-designing 3D anchors. We initialize the box embeddings identically for a tubelet query.

\noindent\textbf{Tubelet attention.} In order to model relations in the tubelet queries, we propose a tubelet-attention (TA) module which contains two self-attention layers (shown in Figure~\ref{fig:overview}). First we have a \textit{spatial self-attention layer} that processes the spatial relations between box query embeddings within a frame \ie $\{q_{1,t},...,q_{N,t}\}$, $t{=}\{1,...,T_\text{out}\}$. The intuition of this layer is that recognizing actions benefits from the interactions between actors, or between actors and objects in the same frame. Next we have our \textit{temporal self-attention layer} that models the correlations between box query embeddings across time within the same tubelet, \ie $\{{q_{i,1}, ..., q_{i,T_\text{out}}}\}$, $i{=}\{1,...,N\}$. This layer facilitates a TubeR query to track actors and generate action tubelets that focus on single actors instead of a fixed area in the frame. TubeR decoder applies the tubelet attention module to tubelet queries $Q$ for generating the tubelet query feature $F_\text{q}\in \mathbb {R}^{N \times T_\text{out} \times C'}$:
\begin{equation}
    F_q = \text{TA}(Q).
    \label{equ: tube_query}
\end{equation}
\noindent\textbf{Decoder.} The decoder contains a tubelet-attention module and a cross-attention (CA) layer which is used to decode the tubelet-specific feature $ F_{\text{tub}}$ from $ F_{\text{en}}$ and $ F_{\text{q}}$:
\begin{equation}
    \text{CA}(F_q, F_{\text{en}})=\text{softmax}(\frac{F_q \times \sigma_{k}(F_\text{en})^T}{\sqrt{C'}})\times \sigma_{v}(F_\text{en}),
    \label{equ: actor_query}
\end{equation}
\begin{equation}
    F_{\text{tub}} = \text{Decoder}(F_q, F_{\text{en}}).
\end{equation}
$F_{\text{tub}} \in \mathbb {R}^{N \times T_\text{out} \times C'} $ is the tubelet specific feature. Note that with temporal pooling, $T_\text{out} < T_\text{in}$, TubeR produces sparse tubelets; For $T_\text{out} {=} T_\text{in}$, TubeR produces dense tubelets.

\subsection{Task-Specific Heads}
\label{sec:head}
The bounding boxes and action classification for each tubelet can be done simultaneously with independent task-specific heads. Such design maximally reduces the computational overheads and makes our system expandable.

\noindent\textbf{Context aware classification head.} The classification can be simply achieved with a linear projection. 
\begin{equation}
    y_{\text{class}} = \text{Linear}_{\text{c}}(F_{\text{tub}}),
    \label{equ:cls}
\end{equation}
where $y_{\text{class}} \in \mathbb{R}^{N \times L}$ denotes the classification score on $L$ possible labels, one for each tubelet. 

 \textit{Short-term context head.} It is known that context is important for understanding sequences~\cite{vaswani2017attention}. We further propose to leverage spatio-temporal video context to help video sequence understanding. We query the action specific feature $F_\text{tub}$ from some context feature $F_\text{context}$ to strengthen $F_\text{tub}$, and get the feature $F_\text{c} \in \mathbb{R}^{N \times C'}$ for the final classification:
\begin{equation}
    F_\text{c} = \text{CA}(\text{Pool}_t(F_\text{tub}), \text{SA}(F_\text{context}))+\text{Pool}_t(F_\text{tub}).
    \label{equ:backbone_query}
\end{equation}
When we set $F_\text{context} {=} F_\text{b}$ for utilizing the short-term context in the backbone feature, we call it short-term context head. A self-attention layer is first applied to $F_\text{context}$, then a cross-attention layer utilizes $F_\text{tub}$ to query from $F_\text{context}$. The $\text{Linear}_{\text{c}}$ is applied to $F_\text{c}$ for final classification.
% More empirical study on where to query the feature are presented in the ablation study.

\textit{Long-term context head.} Inspired by~\cite{wang2018non,zhang2021vidtr, wu2019long} which explore long-range temporal information for video understanding, we propose a long-term context head. To utilize long-range temporal information but under certain memory budget, we adopt a two-stage decoder for long-term context compression as described in~\cite{xu2021long}:
\begin{equation}
    \text{Emb}_{\text{long}} = \text{Decoder}(\text{Emn}_{n1}, \text{Decoder}(\text{Emb}_{n0}, F_{\text{long}}).
\end{equation}
The long-term context $F_\text{long} \in \mathbb{R}^{T_\text{long} \times H'W' \times C'}$ ($T_\text{long} {=} (2w+1)T'$) is a buffer that contains the backbone feature extracted from a long $2w$ adjacent clips concatenated along time. In order to compress the long-term video feature buffer to an embedding $\text{Emb}_\text{long}$ with a lower temporal dimension,
we apply two stacked decoders with two token embedding $\text{Emn}_{n0}$ and $\text{Emn}_{n1}$. Specifically, we first apply a compressed token $\text{Emb}_{n_0}$ ($n_0 < T_\text{long}$) to query important information from $F_\text{long}$ and get an intermediary compressed embedding with temporal dimension $n_0$. Then we further utilize another compressed token $\text{Emb}_{n_1}$ ($n_1 < n_0$) to query from the intermediary compressed embedding and get the final compressed embedding $\text{Emb}_\text{long}$. $\text{Emb}_\text{long}$ contains the long-term video information but with a lower temporal dimension $n_1$. 
% We use two stage decoder for long-term context compression, similar to \cite{xu2021long}; in implementation, we empirically determined $q0=16$ and $q1=32$. \bing{Not sure whether those details can be pushed to implementation details.}
%The $\text{Emb}_\text{long}$ denotes the compressed long-term backbone features with temporal dimension of $q1$. 
Then we adopt a cross-attention layer to $F_\text{b}$ and $\text{Emb}_{\text{long}}$ to generate a long-term context feature $F_\text{lt} \in \mathbb{R}^{T' \times H' \times W' \times C'}$:
\begin{equation}
    F_\text{lt} = \text{CA}(F_\text{b}, \text{Emb}_{\text{long}}),
\end{equation}
we set $F_\text{context} = F_\text{lt}$ in Eq.~\ref{equ:backbone_query} to utilize the long-term context for classification. 

\noindent\textbf{Action switch regression head.}
%\joe{agree, change these subsubtitles to paragraphs and move action switch to the end of this section. I don't think we can change the name now but I think we need to be a bit more clear what it does as others weren't sure.}
%\joe{I feel we are overselling this by calling it action switch regression head. It is predicting if the person is visible and performing the action on the given frame. We used to call this visibility but with UCF it makes sense to call it something else. Even so it feels like by giving it its own section readers be expecting more and think we are over stating novelty. What do others think?}\jj{even it is described simply, but it really helps a lot on UCF. Thus we keep it so far.}
The $T_\text{out}$ bounding boxes in a tubelet are simultaneously regressed with an $\text{FC}$ layer as:
\begin{equation}
    y_{\text{coor}} = \text{Linear}_{\text{b}}(F_{\text{tub}}),
\end{equation}
where $y_{\text{coor}} \in \mathbb{R}^{N \times T_\text{out} \times 4}$, $N$ is the number of action tubelets, and $T_\text{out}$ is the temporal length of an action tubelet. To remove non-action boxes in a tubelet,  we further include an $\text{FC}$ layer for deciding whether a box prediction depicts the actor performing the action(s) of the tubelet, we call action switch. The action switch allows our method to generate action tubelets with a more precise temporal extent. The probabilities of the $T_\text{out}$ predicted boxes in a tubelet being visible are:
\begin{equation}
    y_\text{switch} = \text{Linear}_{\text{s}}(F_\text{tub}),
\end{equation}
where $y_\text{switch} \in \mathbb{R}^{N \times T_\text{out}}$. For each predicted tubelet, each of its $T_\text{out}$ bounding boxes obtain an action switch score. 

\subsection{Losses}
The total loss is a linear combination of four losses:
\begin{equation}
\begin{split}
 \mathcal{L} =\lambda_1\mathcal{L}_{\text{switch}}(y_{\text{switch}},Y_{\text{switch}})
 +\lambda_2\mathcal{L}_{\text{class}}(y_{\text{class}},Y_{\text{class}})\\
 +\lambda_3\mathcal{L}_{\text{box}}(y_{\text{coor}},Y_{\text{coor}})
 +\lambda_4\mathcal{L}_{\text{iou}}(y_{\text{coor}},Y_{\text{coor}}),
\end{split}
\end{equation}
where $y$ is the model output and $Y$ denotes the ground truth. The action switch loss $\mathcal{L}_\text{switch}$ is a binary cross entropy loss. The classification loss $\mathcal{L}_\text{class}$ is a cross entropy loss. The $\mathcal{L}_\text{box}$ and $\mathcal{L}_\text{iou}$ denote the per-frame bounding box matching error. It is noted when $T_\text{out} < T_\text{in}$, the tubelet is sparse and the coordinate ground truth $Y_\text{coor}$ are from the corresponding temporally down-sampled frame sequence. We used the Hungarian matching similar to \cite{carion2020end} and more details can be found in the supplementary. We empirically set the scale parameter as $\lambda_1{=}1,\lambda_2{=}5,\lambda_3{=}2,\lambda_4{=}2$.

%%%%%%%%% 
\section{Experiments}
\label{sec:results}
%=====================================================%
\begin{table*}[t]
    \footnotesize
	\centering
	\subfloat[\textbf{Analysis on tubelet query.} Our tubelet query set design allows for each query to focus on the spatial location of the action on a specific frame.]
	{
	\scalebox{1}{
		\begin{tabularx}{0.3\textwidth}{lrr}
		\toprule
	    &  {\textit{UCF101-24}}  &{\textit{AVA}} \\
			\midrule
           single query   & 48.8  & 26.2\\
           tubelet query set  & 52.9   & 27.4\\
		\bottomrule	
		\end{tabularx}
		}
		\label{tab:tubelet-q}
	}\hfill
	\subfloat[\textbf{Effect of tubelet attention.} With tubelet attention modeling relations within a tubelet and across tubelets improves.]{
	\scalebox{1}{
		\begin{tabularx}{0.27\textwidth}{lrr}
		\toprule
		&  {\textit{UCF101-24}}  &{\textit{AVA}} \\
    			\midrule
               self-attention  & 52.9   & 27.4\\
               tubelet attention   & 53.8  & 27.7\\
    		\bottomrule	
    		\end{tabularx}
		}
		\label{tab:tubelet-a}        
        
    }\hfill
	\subfloat[\textbf{Benefit of action switch.} Action switch produces a more precise temporal extent, which can only be shown by video-mAP.]{
    \scalebox{1}{
		\begin{tabularx}{0.27\textwidth}{lrr}
		\toprule
		& {\textit{UCF101-24}}  &{\textit{AVA}} \\
			\midrule
           w/o switch   & 53.8   & 27.7\\
           w/ switch   & 57.7 & 27.7\\
		\bottomrule	
		\end{tabularx}
        }
        \label{tab:as}
	}\hfill
	\subfloat[\textbf{Effectiveness of short- and long-term context.} The short-term context and long-term context help with performance, more noticeable on AVA.]{
	\scalebox{1}{
		\begin{tabularx}{0.3\textwidth}{lrr}
		\toprule
		& {\textit{UCF101-24}}  &{\textit{AVA}} \\
    			\midrule
               FC head  &  57.8  & 23.4\\
               + short-term context  & 58.4  & 27.7\\
               + long-term context   & -  & 28.8\\
    		\bottomrule	
    		\end{tabularx}
		}
		\label{tab:feature_query}
	}\hfill
	\subfloat[\textbf{Length of input clip.} Longer input video leads to a better performance on both UCF101-24 and AVA.]{
		\scalebox{1}{
		\begin{tabularx}{0.27\textwidth}{lcc} 
    		\toprule
    		& \textit{UCF101-24}  & \textit{AVA} \\
    		\midrule
    		8 &  53.9  & 24.4\\
    		16 & 58.2  & 26.9\\
    		32 & 58.4  & 27.7\\
    		\bottomrule
    	\end{tabularx}}
	\label{tab:ablation_clip_len}
	}\hfill
	\subfloat[\textbf{Long-term context length} analysis on AVA. The right amount of long-term context helps improve frame-mAP on AVA.]{
        \scalebox{1}{
		\begin{tabularx}{0.27\textwidth}{lcrc} 
    		\toprule
    		$w$ & \# of clips  & duration (s) & mAP   \\ 
    		\midrule
    		- & 1  & 2.1  & 27.7\\
    		2 & 5  & 10.6 & 28.4\\
    		3 & 7  & 14.9 & 28.8\\
    		5 & 11 & 23.5 & 28.6\\
    		\bottomrule
    	\end{tabularx}}
	\label{tab:ablation_long_w}
	}
	
	\label{tab:ablation}
    \caption{Ablation studies on UCF101-24 and AVA 2.1. The proposed tubelet query, tubelet attention, the action switch and context-awareness generally improve model performance. The proposed TubeR works well on long clips with shot changes. We report video-mAP@IoU=0.5 for UCF101-24 and frame-mAP@IoU=0.5 for AVA.}
	\vspace{-5mm}
\end{table*}

%--------------------------------------

\subsection{Experimental Setup}

\noindent\textbf{Datasets.} We report experiments on three commonly used video datasets for action detection. \textbf{UCF101-24}~\cite{soomro2012ucf101} is a subset of UCF101. It contains 24 sport classes in 3207 untrimmed videos. %Each video contains a single action class. Multiple instances of the same class can occur in a single video but with different spatial and temporal boundaries. 
We use the revised annotations for UCF101-24 from~\cite{singh2017online} and report the performance on split-1. 
\textbf{JHMDB51-21}~\cite{jhuang2013towards} contains 21 action categories in 928 trimmed videos. We report the average results over all three splits. 
\textbf{AVA}~\cite{gu2018ava} is larger-scale and includes 299 15-minute movies, 235 for training, and the remaining 64 for validating. Box and label annotations are provided on per-second sampled keyframes. We evaluate on AVA with both annotation versions v2.1 and v2.2.

\noindent\textbf{Evaluation criteria.} 
We report the video-mAP at different IoUs on UCF101-24 and JHMDB51-21. As AVA only has keyframe annotations, we report frame-mAP@IoU{=}0.5 following~\cite{gu2018ava} using a single, center-crop inference protocol.

\noindent\textbf{Implementation details.}
We pre-train the backbone on Kinetics-400 \cite{kay2017kinetics}. The encoder and decoder contain 6 blocks on AVA. For the smaller UCF101-24 and JHMDB51-21, we reduce the numbers of blocks to 3 to avoid overfitting. We empirically set the number of tubelet query $N$ to 15. During \textbf{training}, we use the bipartite matching \cite{georgiev2020neural} based on the Hungarian algorithm \cite{kuhn1955hungarian} between predictions and the ground truth. We use the AdamW \cite{loshchilov2017decoupled} optimizer with an initial learning rate $1e{-}5$ for the backbone and $1e{-}4$ for the transformers. We decrease the learning rate 10$\times$ when the validation loss saturates. We set $1e{-}4$ as the weight decay. Scale jittering in the range of (288, 320) and color jittering are used for data augmentation. During \textbf{inference}, we always resize the short edge to 256 and use a single center-crop (1-view). We also tested the horizontal flip trick to create 2-view inference. For fair comparisons with previous methods on UCF101-24 and JHMDB51-21, we also test a two-stream setting with optical flow following~\cite{zhao2019dance}.

\subsection{Ablations}
We perform our ablations on both UCF101-24 and AVA 2.1 to demonstrate the effectiveness of our designs on different evaluation protocols. Only RGB inputs are considered. 
For UCF101-24 with per-frame annotations, we report \textbf{video-mAP} at IoU=0.5. A standard backbone I3D-VGG~\cite{gu2018ava} is utilized and the input length is set to 7 frames if not specified.   
For AVA 2.1 with 1-fps annotation, we only take the model prediction on keyframes and report \textbf{frame-mAP} at IoU=0.5.
We use a CSN-50 backbone~\cite{tran2019video} with a single view evaluation protocol if not specified.

\noindent\textbf{Benefit of tubelet queries.} 
We first show the benefit of the proposed tubelet query sets. Each query set is composed of $T_\text{out}$ per-frame query embeddings (see section \ref{sec:query}), which predict the spatial location of the action on their respective frames. We compare this to using a single query embedding that represents a whole tubelet and must regress $T_\text{out}$ box locations for all frames in the clip. Our results are shown in Table~\ref{tab:tubelet-q}. 
Compared to using a single query embedding, our tubelets query set improves performance by +4.1\% video mAP on UCF101-24, showing that modeling action detection as a sequence-to-sequence task effectively leverages the capabilities of transformer architectures. 

\noindent\textbf{Effect of tubelet attention.} 
In Table~\ref{tab:tubelet-a}, we show using our tubelet attention module helps improve video-\textit{mAP} on UCF101-24 by 0.9\% and 0.3\% on AVA. The tubelet attention saves about 10\% memory ($4,414$MB) compared to the typical self-attention implementation ($5,026$MB) during training (16 frames input with batch size of 1).

\begin{figure}[t]
    \centering
    \includegraphics[width=0.45\textwidth]{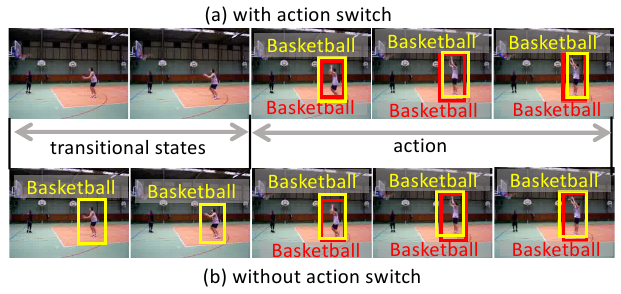}
    \caption{Visualizations of action switch on UCF101-24. Best view in color. The \textcolor{red}{red box and label} represent the ground truth. Yellow indicates our detected tubelets. With the action switch (top row), TubeR avoids misclassification for the transitional states.}
    \label{fig:actionswitch}
    	\vspace{-5mm}
\end{figure}

\noindent\textbf{Benefit of action switch.} 
We report the effectiveness of our action switch head in Table~\ref{tab:as}. On UCF101-24 the action switch increases the video-mAP from 53.8\% to 57.7\% by precisely determining the temporal start and end point of actions. Without action switch, TubeR misclassifies transitional states as actions, like the example shown in Figure~\ref{fig:actionswitch} (bottom row). 
As only the frame-level evaluation can be done on AVA, the advantage of the action switch is not shown by the frame-mAP. Instead, we demonstrate its effect in Figure~\ref{fig:visualization} and Figure~\ref{fig:TubeR_Shot}. The action switch produces tubelets with precise temporal extent for videos with shot changes. 

%====================
%===================
\noindent\textbf{Effect of short and long term context head.} 
We report the impact of our context aware classification head with both short and long-term features in Table \ref{tab:feature_query}. The context head brings a decent performance gain (+4.3\%) on AVA. This is probably because the movie clips in AVA contain shot changes and so the network benefits from seeing the full context of the clip.
On UCF101-24, the videos are usually short and without shot changes. The context does not bring a significant improvement on UCF101-24. 

\noindent\textbf{Length of input clip.} We report results with variable input lengths in Table~\ref{tab:ablation_clip_len}. We compare with input length of 8, 16 and 32 on both UCF101-24 and AVA with CSN-152 as backbone. TubeR is able to handle long video clips as expected. We notice that our performance on UCF101-24 saturates faster than on AVA, probably because UCF101-24 does not contain shot changes that requires longer temporal context for classification. 

\noindent\textbf{Length of long-term context.} This ablation is only conducted on AVA as videos on UCF101-24 are too short to use long-term context. Table \ref{tab:ablation_long_w} shows that the right amount of long-term context helps performance, but overwhelming the amount of long-term context harms performance. This is probably because the long-term feature contains both useful information and noise. The experiments show that about 15s context serves best. Note that the context length varies per dataset, but can be easily determined empirically.

%=====================================================%
%\subsection{Main Results}
\begin{table*}[t]
    %\footnotesize
	\tiny
	\centering
	\resizebox{0.85\linewidth}{!}{

		\begin{tabular}{lcrccccc} 
		    \toprule
			\textbf{Model} & \textbf{Detector}  & \textbf{Input} & \textbf{Backbone} & \textbf{Pre-train} & \textbf{Inference} & \textbf{GFLOPs} &  \textbf{mAP} \\
			\midrule
		    \rowcolor{Gray}
		    %\textbf{Comparison to end-to-end models} & & & & & & & &\\
		    \multicolumn{8}{l}{\textbf{Comparison to end-to-end models}} \\
			I3D \cite{gu2018ava}& \xmark  & 32 $\times$ 2 & I3D-VGG & K400 & 1 view & NA & 14.5\\
			ACRN \cite{sun2018actor}&  \xmark  & 32 $\times$ 2 & S3D-G & K400 & 1 view & NA & 17.4\\			
			STEP \cite{yang2019step}&  \xmark  & 32 $\times$ 2 & I3D-VGG & K400 & 1 view & NA & 18.6\\
			VTr \cite{girdhar2019video}&  \xmark  & 64 $\times$ 1 & I3D-VGG & K400 & 1 view & NA & 24.9\\
			WOO \cite{chen2021watch}&  \xmark  & 8 $\times$ 8 & SF-50 & K400 & 1 view & 142 & 25.2\\
			\textbf{TubeR} & \xmark  & 16 $\times$ 4 & I3D-Res50 & K400 & 1 view & 132& \textbf{26.1}\\
			\textbf{TubeR} & \xmark&  16 $\times$ 4 & I3D-Res101 & K400 & 1 view & 246& \textbf{28.6}\\
			\midrule
			\rowcolor{Gray}
			\multicolumn{8}{l}{\textbf{Comparison to two-stage models}} \\
		    %\textbf{Comparison to two-stage models } & & & & & & & &\\
			Slowfast-50 \cite{feichtenhofer2019slowfast}& F-RCNN & 16 $\times$ 4 & SF-50 & K400 & 1 view & 308 & 24.2\\
			X3D-XL \cite{feichtenhofer2020x3d}& F-RCNN  & 16 $\times$ 5 & X3D-XL & K400 & 1 view& 290 & 26.1\\
			CSN-152*  & F-RCNN  & 32 $\times$ 2 & CSN-152 & IG + K400 & 1 views& 342 & 27.3\\			
			LFB \cite{wu2019long} & F-RCNN  & 32 $\times$ 2 & I3D-101-NL & K400 & 18 views  & NA & 27.7\\		
			ACAR-NET \cite{pan2021actor} & F-RCNN  & 32 $\times$ 2 & SF-50 & K400 & 6 views& NA & 28.3\\			
			\textbf{TubeR} & \xmark  & 32 $\times$ 2 & CSN-50 & K400 & 1 view& 78 & \textbf{28.8}\\
			%\textbf{TubeR}~(ours) & \xmark & \xmark & 32 $\times$ 2 & CSN-50 & K400 & 2 view& 77.9 $\times$ 2 & \textbf{29.1}\\			
			\textbf{TubeR} & \xmark & 32 $\times$ 2 & CSN-152 & IG + K400 & 1 view& 120 & \textbf{31.7}\\
			\midrule
			\rowcolor{Gray}
			%\textbf{Comparison to best  reported results} & & & & & & & &\\
        	\multicolumn{8}{l}{\textbf{Comparison to best reported results}} \\
			WOO \cite{chen2021watch}&  \xmark  & 8 $\times$ 8 & SF-101 & K400+K600 & 1 view & 246 & 28.0\\
			SF-101-NL \cite{feichtenhofer2019slowfast}& F-RCNN & 32 $\times$ 2 & SF-101+NL & K400+K600 & 6 views& 962 & 28.2\\
			ACAR-NET \cite{pan2021actor} & F-RCNN  & 32 $\times$ 2 & SF-101 & K400+K600 & 6 views& NA & 30.0\\
			AIA \cite{tang2020asynchronous}  & F-RCNN & 32 $\times$ 2 & SF-101 & K400+K700 & 18 views & NA & 31.2\\
			\textbf{TubeR} & \xmark  & 32 $\times$ 2 & SF-101 & K400+K700 & 1 view& 240& \textbf{31.6}\\
			\textbf{TubeR} & \xmark  & 32 $\times$ 2 & CSN-152 & IG + K400 & 2 view& 240& \textbf{32.0}\\
			\bottomrule
		\end{tabular}
		}
    	\caption{\textbf{Comparison on AVA v2.1} validation set. Detector shows if additional detector is required; * denotes the results we tested. IG denotes the IG-65M dataset, SF denotes the slowfast network. The FLOPs for two-stage models are the sum of Faster RCNN-R101-FPN FLOPs (246 GFLOPs \cite{carion2020end}) plus classifier FLOPs multiplied by view number. TubeR performs more effectively and efficiently.}
    		\label{tab:ava_sota}
	\vspace{-3mm}
\end{table*}

\begin{table}[t]
	\centering
	\resizebox{1\linewidth}{!}{
		\begin{tabular}{lcccc}

		    \toprule
			\textbf{Model} & \textbf{backbone} & \textbf{pre-train} & \textbf{inference} &  \textbf{mAP} \\
			\midrule
			\rowcolor{Gray}
			\multicolumn{5}{l}{\textbf{Single-view}} \\
            X3D-XL \cite{feichtenhofer2020x3d} & X3D-XL & K600+ K400 & 1 view & 27.4\\
            CSN-152 \cite{Zhang_2021_CVPR} & CSN-152 & IG + K400 & 1 view & 27.9\\            
            WOO \cite{chen2021watch}  & SF-101 & K600+ K400 & 1 view & 28.3\\
            M-ViT-B-24 \cite{fan2021multiscale}  & MViT-B-24 & K600+ K400 & 1 view & 28.7\\
            \textbf{TubeR} & CSN-50 & IG + K400 & 1 view & 29.2\\
            \textbf{TubeR} & CSN-152 & IG + K400 & 1 view & 33.4\\
            \midrule
            \rowcolor{Gray}
			\multicolumn{5}{l}{\textbf{Multi-view}} \\
            SlowFast-101 \cite{feichtenhofer2019slowfast} & SF-101 & K600+ K400 & 6 views & 29.8\\
            ACAR-Net \cite{pan2021actor} & SF-101 & K700+ K400 & 6 views & 33.3\\
            AIA (obj) \cite{tang2020asynchronous} & SF-101 & K700+ K400 & 18 views & 32.2\\
            %\midrule
            \textbf{TubeR} & CSN-152 & IG + K400 & 2 views & \textbf{33.6}\\
			\bottomrule
		\end{tabular}
		}
		
    	\caption{\textbf{Comparison on AVA v2.2} validation set. IG denotes the IG-65M, SF denotes the slowfast. TubeR achieves the best result.}
    	
    	\label{tab:ava_v22_sota}
	\vspace{-5mm}
\end{table}

\subsection{Frame-Level State-of-the-Art}

\noindent\textbf{AVA 2.1 Comparison.}
We first compare our results with previously proposed methods on AVA 2.1 in Table \ref{tab:ava_sota}. Compared to previous end-to-end models, with comparable backbone (I3D-Res50) and the same inference protocol, the proposed TubeR outperforms all. TubeR outperforms the most recent end-to-end works WOO \cite{chen2021watch} by 0.9\% and VTr \cite{girdhar2019video} by 1.2\%. This demonstrates the effectiveness of our designs. 

Compared to previous work using an offline person detector, the proposed TubeR is also more effective under the same inference protocols. This is because TubeR generates tubelet-specific features without assumptions on location, while the two-stage methods have to assume the actions occur at a fixed location. It is also worth mentioning that the TubeR with CSN backbones outperforms the two-stage model with the same backbone by +4.4\%, demonstrating that the gain is not from the backbone but our TubeR design. TubeR even outperforms the methods with multi-view augmentations (horizontal flip, multiple spatial crops and multi-scale).
TubeR is also considerably faster than previous models, we have attempted to collect the reported FLOPs from previous works (Table \ref{tab:ava_sota}). Our TubeR has 8\% fewer FLOPs than the most recently published end-to-end model \cite{chen2021watch} with higher accuracy. 
Tuber is also $4\times$ more efficient than the two-stage model \cite{feichtenhofer2019slowfast} with noticeable performance gain. Thanks to our sequence-to-sequence design, the heavy backbone is shared and we do not need temporal iteration for tubelet regression.

We finally present the highest number reported in the literature, regardless of the inference protocol, pre-training dataset and additional information used. TubeR still achieves the best performance, even better than the model using additional object bounding-boxes as input \cite{tang2020asynchronous}.The results show that the proposed sequence-to-sequence model with tubelet specific feature is a promising direction for action detection.  

\noindent\textbf{AVA 2.2 Comparison.}
The results are shown in Table \ref{tab:ava_v22_sota}.
Under the same single-view protocol, TubeR is considerably better than previous methods, including the most recent work with an end-to-end design (WOO \cite{chen2021watch} +5.1\%) and the two-stage work with strong backbones (MViT \cite{fan2021multiscale} +4.7\%). A fair comparison between TubeR and a two-stage model~\cite{Zhang_2021_CVPR} with the same backbone CSN-152, shows TubeR gains +5.5\% frame-mAP. It demonstrates TubeR's superior performance comes from our design rather than the backbone.
%The TubeR's superior performance comes from our design rather than the backbone, since two-stage model with same CSN-152 backbone barely achieves comparable performance to WOO and M-ViT \cite{chen2021watch,fan2021multiscale}.

%-----------------
\noindent\textbf{UCF101-24 Comparison.}
We also compare TubeR with the state-of-the-art using frame-mAP@IoU=0.5 on UCF101-24 (see the first column with numbers in Table~\ref{tab:sota}). Compared to existing methods, TubeR acquires better results with comparable backbones, for both RGB-stream and two-stream settings. Further with a CSN-152 backbone, TubeR gets 83.2 frame-mAP, even better than two-stream methods. Though TubeR targets on tubelet-level detection, it performs well on frame-level evaluation on both AVA and UCF101-24.

%=====================================================%

\subsection{Video-Level State-of-the-Art}

\begin{table}[t!]
	\centering
	\LARGE
	\resizebox{1.\linewidth}{!}{
		\begin{tabular}{lccccccc}
		    \toprule
	  & & \multicolumn{4}{c}{\textit{\textbf{UCF101-24}}} & \multicolumn{2}{c}{\textit{\textbf{JHMDB51-21}}}   \\ 
		\cmidrule(lr){3-6} \cmidrule(lr){7-8} 
		& \textbf{Backbone}& {{f-mAP}} & {{0.20}} & {{0.50}} & {{0.50:0.95}}  & {{0.20}}& {{0.50}} \\
		\midrule
%SINGLE
			\rowcolor{Gray}
        	\multicolumn{8}{l}{\textbf{RGB-stream}} \\
		   MOC~\cite{li2020actions}& DLA34 &72.1& 78.2 & 50.7 & 26.2  & - & - \\
		   \textbf{TubeR}& Res50 &79.5& 81.2 & 55.1 & 28.1   & - & - \\
		   T-CNN~\cite{hou2017tube} & C3D & 41.4 & 47.1 & - & -  & 78.4 & 76.9\\
		   \textbf{TubeR}& I3D & 80.1 & 82.8 & 57.7 & 28.6   & 79.7 & 78.3 \\
		   \textbf{TubeR}& CSN-152 &\textbf{83.2}& \textbf{83.3} & \textbf{58.4} & \textbf{28.9}   & \textbf{87.4} & \textbf{82.3} \\
\midrule
			\rowcolor{Gray}
        	\multicolumn{8}{l}{\textbf{Two-stream}} \\
           TacNet~\cite{song2019tacnet} &VGG &72.1& 77.5&52.9 &24.1 & -  & - \\
		   2in1~\cite{zhao2019dance}  & VGG && 78.5 & 50.3  & 24.5 & -& 74.7 \\
		   ACT~\cite{kalogeiton2017action} & VGG &67.1& 77.2 & 51.4 & 25.0 & 74.2 & 73.7\\
		   MOC~\cite{li2020actions}  &DLA34 &78.0& 82.8 & 53.8& 28.3  & 77.3 & 77.2 \\
           STEP~\cite{yang2019step} &I3D &75.0& 76.6 &- & -  & - & -\\
		   I3D~\cite{gu2018ava}&I3D & 76.3 &-& 59.9 & -& - & 78.6\\
		   *CFAD~\cite{li2020cfad}&I3D & 72.5 &81.6& \textbf{64.6} & 26.7& \textbf{86.8} & \textbf{85.3}\\
		   \textbf{TubeR}& I3D &\textbf{81.3}& \textbf{85.3} & 60.2 & \textbf{29.7}   & 81.8 & 80.7 \\
			\bottomrule

		\end{tabular}
}
	\caption{
	\textbf{Comparison on UCF101-24 and JHMDB51-21} with video-\textit{mAP}. TubeR achieves better results compared to most state-of-arts. f-mAP denotes the frame mAP@IoU=0.5. *CFAD is pretrained on K600 but others on K400.}
	\label{tab:sota}
\vspace{-5mm}
\end{table}

We also compare TubeR with various settings to  state-of-the-art reporting video-mAP on UCF101-24 and JHMDB51-21 in Table~\ref{tab:sota}. For fair comparisons, TubeR with a 2D backbone gains +4.4\% video-mAP@IoU=0.5 compared to the recent state-of-the-art~\cite{li2020actions} on UCF101-24 without using optical flow, which demonstrates that TubeR learning tubelet queries is more effective compared to using positional hypotheses. Compared to TacNet~\cite{song2019tacnet} which proposes a transition-aware context network to distinguish transitional states, TubeR with action switch performs better even with a one-stream setting. When incorporating optical flow inputs, the TubeR with I3D further boosts the video-level results. It is noted that TubeR pretrained on K400 even outperforms CFAD pretrained on K600 on some metrics. We test TubeR inference speed on UCF101-24 by following CFAD. To directly generate a tubelet without an offline linker, TubeR runs at 156 fps. Faster than CFAD (130fps) and most existing SOTA methods (40-53 fps). The result illustrates our design is effective and efficient for video-level action detection. 

%=====================================================%

\subsection{Visualization}
\begin{figure}[t]
    \centering
    \includegraphics[width=0.9\columnwidth]{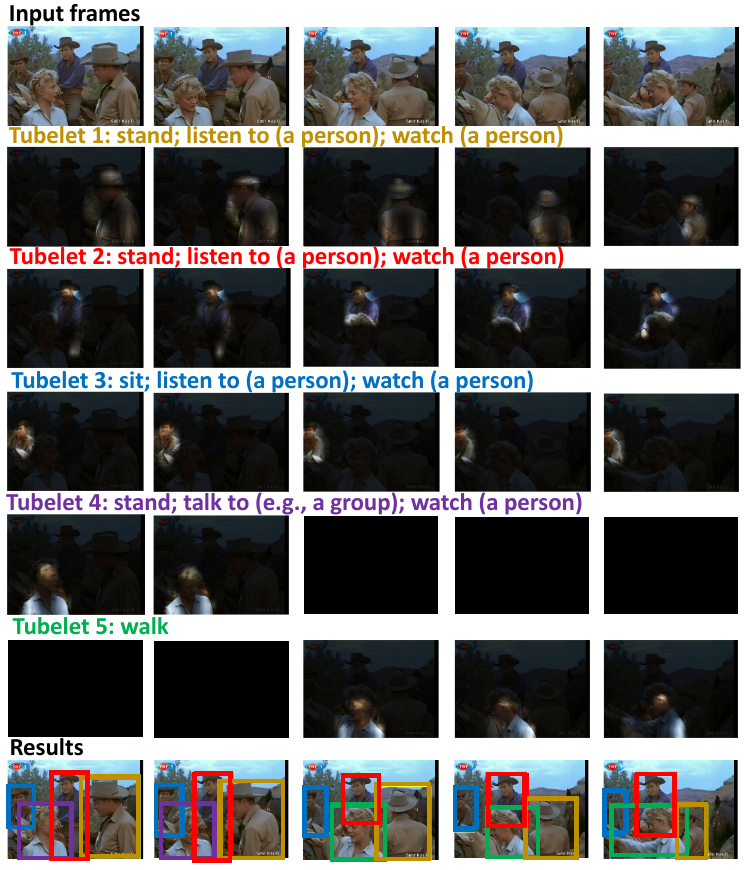}
    \caption{Visualization of tubelet specific feature with attention rollout. Each tubelet covers a separated action instance. Best viewed in color.}
    \label{fig:visualization}
\vspace{-3mm}
\end{figure}

We first provide visualizations (Figure \ref{fig:visualization}) of the tubelet-specific features by overlaying the tubelet-specific feature activation over the input frames using attention rollout \cite{abnar2020quantifying}. The example in Figure \ref{fig:visualization} is challenging as it contains multiple people and concurrent actions. The visualization show that: 1. Our proposed TubeR is able to generate highly discriminative tubelet-specific features. Different actions in this case are clearly separated in different tubelets. 2. Our action switch works as expected and initiates/cuts the tubelets when the action starts/stops. 3. Our TubeR generalizes well to scale changes (the brown tubelet). 4. The generated tubelets are tightly associated with tubelet specific feature as expected. 

\begin{figure}[t]
    \centering
    \includegraphics[width=0.9\columnwidth]{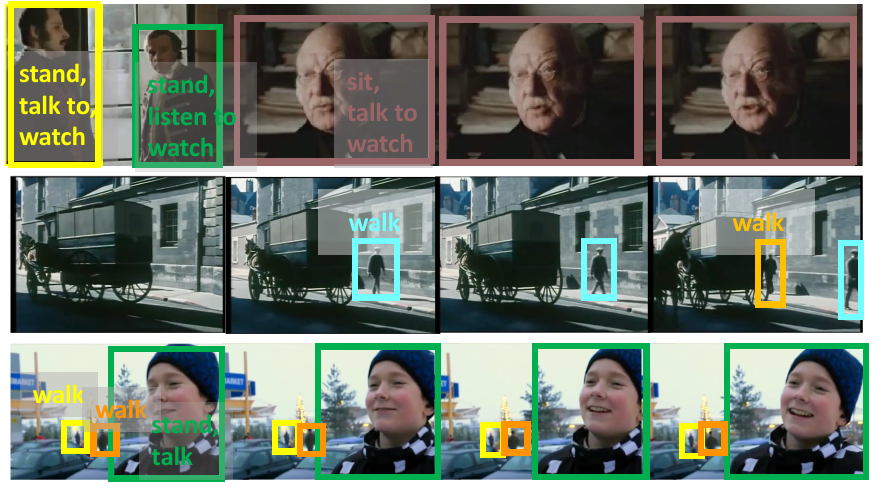}
    \caption{Results visualization, with different colors to label different tubelets. Each action tubelet contains its action labels and boxes per frame. We only show the action labels on the first frame of an action tubelet. Some challenging cases are shown. Top: shot changes; Middle: actors moving with distance; Bottom: multiple actors with small and large scales. Best viewed in color.}
    \label{fig:TubeR_Shot}
\vspace{-5mm}
\end{figure}

We further show our TubeR performs well in various scenarios. TubeR works well on videos with shot changes (Figure \ref{fig:TubeR_Shot} top); TubeR is able to detect an actor moving with distance (Figure \ref{fig:TubeR_Shot} middle); and TubeR is robust to action detection even for small people (Figure \ref{fig:TubeR_Shot} bottom).

%%%%%%%%% 
% \section{Visualization}
% \label{sec:discussion}
% \input{tex/5_vis}

%%%%%%%%% 
\section{Discussion and Conclusion}
\label{sec:conclusion}
\noindent\textbf{Limitations.} 
Although proposed for long videos, we noticed two potential limitations that stop us from feeding in very long videos in one shot. 

\noindent1. We observe that 90\% of computation (FLOPs) and 67\% of memory usage was used by our 3D backbone. This heavy backbone restricts us from applying TubeR on long videos. Recent works show that transformer encoders can be used for video embedding \cite{zhang2021vidtr,fan2021multiscale,arnab2021vivit} and are less memory and computationally hungry. We will explore these transformer based embeddings in future work.

\noindent2. If we were to process a long video in one pass we'd need enough queries to cover the maximum number of different actions per-person in that video. This would likely require a large number of queries which would cause memeory issues in our self attention layers. A possible solution is to generate person tubelets, instead of action tubelets, so that we do not need to split tubelets when a new action happens. Then we would only need a query for each person instance.
%Recall the tubelet query: $Q_{\text{tub}} \in \mathbb{R}^{N \times T_\text{out} \times C'}$, where $N$ denotes the maximum actions can be detected. We have to set N  to a large number (e.g. 1000) for the long video or video with a large number of people. This lead to long query embedding and potentially cause memory issue. A possible solution is to generate person tubelet, instead of action tubelet, so that we don't need to split tubelets when a new action happens. 

\noindent\textbf{Potential negative impact.} There are real-world applications of action detection technology such as patient or elderly health monitoring, public safety, Augmented/Virtual Reality, and collaborative robots. However, there could be unintended usages and we advocate responsible usage and complying with applicable laws and regulations.

\noindent\textbf{Conclusion.} This paper introduces TubeR, a unified solution for spatio-temporal video action detection in a sequence-to-sequence manner.
Our design of tubelet-specific features allows TubeR to generate tubelets (a set of linked bounding boxes) with action predictions for each of the tubelets.
TubeR does not rely on positional hypotheses and therefore scales well to longer video clips.
TubeR achieves state-of-the-art performance and better efficiency compared to previous works. 
%TubeR establishes a new paradigm for spatio-temporal action detection, 
% We will further extend the TubeR idea to more complicated tasks such as video segmentation, pose estimation and tracking. 
% \paragraph{Action Tube vs. Actor Tube}

%%%%%%%%% REFERENCES
{\small
\bibliographystyle{ieee_fullname}
\bibliography{ADTR}
}

\end{document}